# Weakly supervised object detection using pseudo-strong labels

Ke Yang [a, b*], Dongsheng Li [a, b], Yong Dou[a, b], Shaohe Lv[a, b], Qiang Wang[a, b]

[a] National Laboratory for Parallel and Distributed Processing,

National University of Defense Technology, Changsha 410073, China

[b] College of Computer, National University of Defense Technology,

Changsha 410073, China

**Abstract**

Object detection is an import task of computer vision. A variety of methods have been proposed, but methods using the weak labels still do not have a satisfactory result. In this paper, we propose a new framework that using the weakly supervised method's output as the pseudo-strong labels to train a strongly supervised model. One weakly supervised method is treated as black-box to generate class-specific bounding boxes on train dataset. A de-noise method is then applied to the noisy bounding boxes. Then the de-noised pseudo-strong labels are used to train a strongly object detection network. The whole framework is still weakly supervised because the entire process only uses the image-level labels. The experiment results on PASCAL VOC 2007 prove the validity of our framework, and we get result 43.4% on mean average precision compared to 39.5% of the previous best result and 34.5% of the initial method, respectively. And this frame work is simple and distinct, and is promising to be applied to other method easily.

---
[*] Corresponding author. E-mail address: bityangke@163.com (Ke Yang)

**1. Introduction and Related Works**

In recent years, deep learning networks have dominate the computer vision field, such as object detection [1, 2, 3], image classification [4], semantic segmentation [5] and human pose estimation [6]. Convolutional Neural Networks (CNN) is the core factor of success due to its ability of learning generic visual features that can be transferred to many domain. Object detection is one of the main tasks in image understanding. A series of works have been proposed to solve this problem, and have gotten promising results. However, Most of the works are using strongly supervised method. In this work, we focus on the weakly supervised object detection task, of which the results are still very poor. The research on supervised image recognition is necessary due to a few reasons. First, the training procedure of CNN consumes large amounts of labeled data. Second, complex annotation in tasks such object detection and image segmentation spends a lot of time and manual effort. At the same time, weakly supervised detection networks only need image-level labels, which are easily annotated. Image-level labels identify the absent or not of the object class.

In recent years, more and more supervised object detection methods have been proposed. There are two promising stream [8]. The first are multiple instance learning (MIL) methods [7]. MIL alternates between selecting object regions of interest and estimating an appearance model of the object using the selected regions. The method has the tendency to get stuck in local optima. The second is the one with CNN [8]. Method of this category is inspired by the facts that without location information a pre-trained CNNs for image classification can learn the information of objects and object parts [9].

Our method fall into the second category: CNN based method. We propose a new framework that using output of Weakly Supervised Deep Detection Networks (WSDDN) [8] on train dataset as the pseudo-strong labels to train a strongly supervised model, such as [2, 3]. The experiment result on PASCAL VOC 2007 dataset [10] is 43.4% on mAP with an improvement of 8.9% compared to initial method, and is 3.9% over the state-of-the-art result.

At a deeper level, a purpose of this work is to verify the validity of the existing strong supervised detection networks such as Fast R-CNN [2] and Faster R-CNN [3] when the labels are noisy. The accurate annotation of object's bounding boxes is difficult, for example, when the object is blocked, the annotation will be difficult to give and includes the annotator's subjectiveness. At the same time, the accurate annotation costs lots of efforts. The annotations may be noisy. It is meaningful to see how the noise annotations will influence the performance of the state-of-the-art strong supervised detection network, in other words, the ability of these networks tolerate noise. Then we take consideration weakly supervised learning algorithms generating labels automatically. Taking object detection as example, without training with bounding boxes information, the detection output of supervised network must be noisy. We use the detection results of weakly supervised network on train dataset as pseudo-strong labels to train a strong supervised network. Then it is interested to see how the performance changes. The whole framework is still *weakly supervised* because the entire process only uses the image-level labels.

## 2. Proposed Framework

This section describes the proposed framework. First, we introduce the networks we use to generate the pseudo-strong labels. Second, we introduce the de-noising process to the noisy pseudo-strong labels. Third, we introduce the strongly supervised networks we used and the training settings.

2.1 Weakly supervised detection network

To generate the train dateset's pseudo-labels with bounding boxes, we use the WSDDN's [8] detection output on PASCAL VOC 2007's [10] train + val image set. We use the following settings: EdgeBoxes [11] for candidate regions generating, using the small pre-trained CNN for classification and using the spatial regulariser. For the detection output threshold, we use the default setting that is 0.1.

2.2 Pseudo-labels Filter

Due to the knowledge gap between training and testing of weakly supervised detection network, the detection results on train dataset are noisy. At the same time, we have the image-level labels, we can filter the detection results whose class does not appear in the image. This simple process can improve the result to a certain degree. And we are pursuing more powerful denoising method.

2.3 Supervised Detection Network

To make it convincing and taking the time consumption into consideration at the same time, we choose two state-of-the-art strong supervised detection network [2, 3]. For each network, we just use their default settings, for pre-trained CNN model we use VGG16 [12]. The annotation for training are generated from first two procedure.

## 3. Experiments Results

3.1 Dataset

We use the most widely used dataset PASCAL VOC 2007 [10] for the convenience to compare our results with state-of-the-art results. PASCAL VOC 2007 dataset includes 2501 training, 2510 validation, and 5011 testing images for 20 object categories containing both classification and bounding box annotations. We adopt the suggested splits for training, validation, and testing sets. We train the model on training and validation sets, and evaluate on testing set. As for performance evaluation, we adopt the most widely used standard PASCAL VOC protocol and report the average precision (AP) at 50% intersection-over-union (IoU) of the detected boxes with the ground truth ones.

3.2 Detection Results

The detection results on 20 categories and mean AP (mAP) are shown in Table 1. We compare our results with state-of-the-art method, namely, WSDDN [8] and WSL [9]. Note, we report the best results in [9] which ensemble three networks' outputs to make final detection decision. Initial network of our framework is the small model of [8], and the performance is significantly poorer than the ensemble one as we show in Table 1.

| Method | aero | bike | bird | boat | bottle | bus | car | cat | chair | cow | table | dog | horse | mbike | person | plant | sheep | sofa | train | tv | Avg. |
|---|---|---|---|---|---|---|---|---|---|---|---|---|---|---|---|---|---|---|---|---|---|
| WSL[9] | **54.5** | 47.4 | 41.3 | 20.8 | **17.7** | 51.9 | 63.5 | 46.1 | **21.8** | **57.1** | 22.1 | 34.4 | 50.5 | 61.8 | 16.2 | **29.9** | 40.7 | 15.9 | 55.3 | 40..2 | 39.5 |
| WSDDN Ensemble[8] | 46.4 | 58.3 | 35.5 | **25.9** | 14.0 | **66.7** | 53.0 | 39.2 | 8.9 | 41.8 | 26.6 | 38.6 | 44.7 | 59.0 | 10.8 | 17.3 | 40.7 | **49.6** | **56.9** | 50.8 | 39.3 |
| WSDDN Small [8] | 42.9 | 56.0 | 32.0 | 17.6 | 10.2 | 61.8 | 50.2 | 29 | 3.8 | 36.2 | 18.5 | 31.1 | 45.8 | 54.5 | 10.2 | 15.4 | 36.3 | 45.2 | 50.1 | 43.8 | 34.5 |
| Our WSDDN +Faster(noise) | 42.3 | 57.7 | 33.7 | 8.8 | 10.7 | 54.1 | 56.0 | 42.3 | 5.7 | 32.9 | 32.6 | 27.5 | 52.0 | 57.2 | 19.3 | 23.9 | 42.3 | 40.5 | 43.8 | 42.0 | 36.3 |
| Our WSDDN +Faster | 45.5 | 61.1 | 39.0 | 14.7 | 12.1 | 54.3 | 57.2 | 51.9 | 2.2 | 39.6 | 36.8 | 38.2 | **53.4** | 61.3 | 21.0 | 19.9 | 40.1 | 46.9 | 48.2 | 42.4 | 39.3 |
| Our WSDDN +Fast(noise) | 51.9 | 65.6 | 39.5 | 17.6 | 11.4 | 56.3 | **64.5** | 55.2 | 2.5 | 46.5 | 39.4 | 32.1 | 51.9 | **66.8** | **22.3** | 25.3 | 46.3 | 45.7 | 48.8 | 46.2 | 41.8 |
| Our WSDDN +Fast | 51.5 | **66.1** | **45.5** | 19.4 | 11.0 | 56.6 | **64.5** | **57.3** | 3.0 | 51.1 | **42.7** | **41.8** | 51.9 | 64.8 | 21.6 | 27.4 | **46.4** | 46.1 | 47.8 | **51.4** | **43.4** |

**Table 1 .** PASCAL VOC 2007 test detection average precision (%). Comparison of our framework on PASCAL VOC 2007 to the state-of-the-art in terms of AP. And improvement of our framework on PASCAL VOC 2007 when de-noising the pseudo-labels

The framework using the Fast R-CNN results in 43.4% on mAP compared to 39.5% of the previous best results and 34.5% of the initial method, respectively.

The results of framework using the Faster R-CNN is 39.3% and comparable to state-of-the-art methods.

The reasons why performance of framework Faster R-CNN is worse are as below. The pseudo-labels generating from the initial model is still very noisy although we have implemented a de-noising procedure. The Region Proposal Network (RPN) — object proposal generating network of Faster R-CNN is also trained on the pseudo-labels, the error accumulate much more than framework using Fast R-CNN which has the separate object proposal parts, such as Selective Search (SS).

3.3 Does de-noising matter?

To show the performance improvement from the de-nosing procedure, we show the results in Table 1. The improvement is significant. The results show the de-noising indeed contributes to performance. And a more powerful de-noising method worth being studied.

**4. Conclusion**

In this paper, we explore a weakly supervised object detection framework. One weakly supervised method is treated as black-box to generate class-specific bounding boxes on train dataset. Then A de-noise method is then applied to the noisy bounding boxes. Strongly supervised object detection networks are trained on the de-noised pseudo-strong labels. The whole framework is still weakly supervised because the entire process only uses the image-level labels. The experiment results show that the performance improves significantly from the initial network and exceed state-of-the-art method to a large extent. And our experiment results also suggest that the weakly supervised detection methods still have long way to go and the pseudo-labels generating automatically from weakly supervised methods are still not enough to train the state-of-the-art strong supervised detection networks.

**Acknowledgements**

We thank Hakan Bilen for the helpful answer. This work was supported by the National Natural Science Foundation of China under Grants 61125201, U1435219, and 61572515.

Weakly supervised object detection using pseudo-strong labels